\documentclass[sigconf]{acmart}

\usepackage{booktabs} 
\usepackage{subcaption}
\usepackage[countmax]{subfloat}
\usepackage{caption} 
\usepackage{tabularx}
\usepackage{multirow}
\usepackage{float}
\usepackage{graphicx}
\usepackage{mathtools}
\usepackage{textcomp}

\setcopyright{none}

\begin{document}
\title[Socioeconomic Dependencies of Linguistic Patterns in Twitter: A Multivariate Analysis]{Socioeconomic Dependencies of Linguistic Patterns\\ in Twitter: A Multivariate Analysis}\titlenote{supported by the SoSweet ANR project (ANR-15-CE38-0011-03).}

\author{Jacob Levy Abitbol}
\affiliation{%
  \institution{Univ Lyon, ENS de Lyon, Inria, CNRS, UCB Lyon 1, LIP UMR 5668, IXXI}
  \city{Lyon} 
  \state{France} 
}
\email{jacob.levy-abitbol@ens-lyon.fr}

\author{M\'arton Karsai}
\affiliation{%
  \institution{Univ Lyon, ENS de Lyon, Inria, CNRS, UCB Lyon 1, LIP UMR 5668, IXXI}
  \city{Lyon} 
  \state{France} 
}
\email{marton.karsai@ens-lyon.fr}

\author{Jean-Philippe Magu\'e}
\affiliation{%
  \institution{ENS de Lyon, ICAR
UMR 5191, CNRS}
  \city{Lyon} 
  \state{France} 
}
\email{jean-philippe.mague@ens-lyon.fr}

\author{Jean-Pierre Chevrot}
\affiliation{%
  \institution{Lidilem, University of Grenoble Alpes}
  \city{Grenoble} 
  \state{France} 
}
\email{jean-pierre.chevrot@u-grenoble3.fr}

\author{Eric Fleury}
\affiliation{%
  \institution{Univ Lyon, ENS de Lyon, Inria, CNRS, UCB Lyon 1, LIP UMR 5668, IXXI}
  \city{Lyon} 
  \state{France} 
}
\email{eric.fleury@ens-lyon.fr}
\fancyhead{}  
\renewcommand{\shortauthors}{J. Levy Abitbol, M. Karsai, J.-P. Magu\'e, J.-P. Chevrot, E. Fleury}

\begin{abstract}
Our usage of language is not solely reliant on cognition but is arguably determined by myriad external factors leading to a global variability of linguistic patterns. This issue, which lies at the core of sociolinguistics and is backed by many small-scale studies on face-to-face communication, is addressed here by constructing a dataset combining the largest French Twitter corpus to date with detailed socioeconomic maps obtained from national census in France. We show how key linguistic variables measured in individual Twitter streams depend on factors like socioeconomic status, location, time, and the social network of individuals. We found that \emph{(i)} people of higher socioeconomic status, active to a greater degree during the daytime, use a more standard language; \emph{(ii)} the southern part of the country is more prone to use more standard language than the northern one, while locally the used variety or dialect is determined by the spatial distribution of socioeconomic status; and \emph{(iii)} individuals connected in the social network are closer linguistically than disconnected ones, even after the effects of status homophily have been removed. Our results  inform sociolinguistic theory and may inspire novel learning methods for the inference of socioeconomic status of people from the way they tweet.
\end{abstract}

%
%


\copyrightyear{2018}
\acmYear{2018} 
\setcopyright{iw3c2w3}
\acmConference[WWW 2018]{The 2018 Web Conference}{April 23--27, 2018}{Lyon, France}
\acmBooktitle{WWW 2018: The 2018 Web Conference, April 23--27, 2018, Lyon, France}
\acmPrice{}
\acmDOI{10.1145/3178876.3186011}
\acmISBN{978-1-4503-5639-8/18/04} 

\maketitle

\section{Introduction}

Communication is highly variable and this variability contributes to language change and fulfills social functions. Analyzing and modeling data from social media allows the high-resolution and long-term follow-up of large samples of speakers, whose social links and utterances are automatically collected. This empirical basis and long-standing collaboration between computer and social scientists could dramatically extend our understanding of the links between language variation, language change, and society.

Languages and communication systems of several animal species vary in time, geographical space, and along social dimensions. Varieties are shared by individuals frequenting the same space or belonging to the same group. The use of vocal variants is flexible. It changes with the context and the communication partner and functions as "social passwords" indicating which individual is a member of the local group~\cite{henry_dialects_2015}. Similar patterns can be found in human languages if one considers them as evolving and dynamical systems that are made of several social or regional varieties, overlapping or nested into each other. Their emergence and evolution result from their internal dynamics, contact with each other, and link formation within the social organization, which itself is evolving, composite and multi-layered~\cite{kretzschmar_language_2010, laks_why_2013}.

The strong tendency of communication systems to vary, diversify and evolve seems to contradict their basic function: allowing mutual intelligibility within large communities over time. Language variation is not counter adaptive. Rather, subtle differences in the way others speak provide critical cues helping children and adults to organize the social world~\cite{kinzler_native_2007}. Linguistic variability contributes to the construction of social identity, definition of boundaries between social groups and the production of social norms and hierarchies. 

Sociolinguistics has traditionally carried out research on the quantitative analysis of the so-called linguistic variables, i.e. points of the linguistic system which enable speakers to say the same thing in different ways, with these variants being \emph{"identical in reference or truth value, but opposed in their social [...] significance"}~\cite{labov_sociolinguistic_1972}. Such variables have been described in many languages: variable pronunciation of -ing as [in] instead of [i\ng{}] in English (\emph{playing} pronounced \emph{playin'}); optional realization of the first part of the French negation (\emph{je (ne) fume pas}, "I do not smoke"); optional realization of the plural ending of verb in Brazilian Portuguese (eles disse(ram), "they said").  For decades, sociolinguistic studies have showed that hearing certain variants triggers social stereotypes~\cite{campbell-kibler_new_2010}. The so-called standard variants (\emph{e.g.} [i\ng{}], realization of negative ne and plural -ram) are associated with social prestige, high education, professional ambition and effectiveness. They are more often produced in more formal situation. Non-standard variants are linked to social skills, solidarity and loyalty towards the local group, and they are produced more frequently in less formal situation.

It is therefore reasonable to say that the sociolinguistic task can benefit from the rapid development of computational social science~\cite{lazer_life_2009}: the similarity of the online communication and face-to-face interaction~\cite{hert_quasi-oralite_1999} ensures the validity of the comparison with previous works. In this context, the nascent field of computational sociolinguistics found the digital counterparts of the sociolinguistic patterns already observed in spoken interaction. However a closer collaboration between computer scientists and sociolinguists is needed to meet the challenges facing the field~\cite{Nguyen:2016}:
\begin{itemize}
\item{Going beyond lexical variation (standard or non-standard usage of words) and English language}
\item{Extending the focus to factors unexplored in digital communication such as social class}
\item{Using the social sciences as a source of methodological inspiration for controlling for multiple factors instead of focusing on one factor as in the field of computational sociolinguistics}
\item{Emphasizing the interpretability of the models and the insights for sociolinguistic theory.}
\end{itemize}
The present work meets most of these challenges. 
It constructs the largest dataset of French tweets enriched with census sociodemographic information existent to date to the best of our knowledge.  From this dataset, we observed variation of two grammatical cues and an index of vocabulary size in users located in France. We study how the linguistic cues correlated with three features  reflective of the socioeconomic status of the users, their most representative location and their daily periods of activity on Twitter. We also observed whether connected people are more linguistically alike than disconnected ones. Multivariate analysis shows strong correlations between linguistic cues and socioeconomic status as well as a broad spatial pattern never observed before, with more standard language variants and lexical diversity in the southern part of the country. Moreover, we found an unexpected daily cyclic evolution of the frequency of standard variants. Further analysis revealed that the observed cycle arose from the ever changing average economic status of the population of users present in Twitter through the day. Finally, we were able to establish that linguistic similarity between connected people does arises partially but not uniquely due to status homophily (users with similar socioeconomic status are linguistically similar and tend to connect). Its emergence is also due to other effects potentially including other types of homophilic correlations or influence disseminated over links of the social network. Beyond we verify the presence of status homophily in the Twitter social network our results may inform novel methods to infer socioeconomic status of people from the way they use language. Furthermore, our work, rooted within the web content analysis line of research \cite{Hovy}, extends the usual focus on aggregated textual features (like document frequency metrics or embedding methods) to specific linguistic markers, thus enabling sociolinguistics knowledge to inform the data collection process.

\section{Related Work}  

For decades, sociolinguistic studies have repeatedly shown that speakers vary the way they talk depending on several factors. These studies have usually been limited to the analysis of small scale datasets, often obtained by surveying a set of individuals, or by direct observation after placing them in a controlled experimental setting. In spite of the volume of data collected generally, these studies have consistently shown the link between linguistic variation and social factors~\cite{chambers,Labov1966Social}. 

Recently, the advent of social media and publicly available communication platforms has opened up a new gate to access individual information at a massive scale. Among all available social platforms, Twitter has been regarded as the choice by default, namely thanks to the intrinsic nature of communications taking place through it and the existence of data providers that are able to supply researchers with the volume of data they require. Work previously done on demographic variation is now relying increasingly on corpora from this social media platform as evidenced by the myriad of results showing that this resource reflects not only  morpholexical variation of spoken language but also geographical~\cite{Eisenstein, Pavalanathan}. 

Although the value of this kind of platform for linguistic analysis has been more than proven, the question remains on how previous sociolinguistic results scale up to the sheer amount of data within reach and how can the latter enrich the former. To do so, numerous studies have focused on enhancing the data emanating from Twitter itself. Indeed, one of the core limitations of Twitter is the lack of reliable sociodemographic information about the sampled users as usually data fields such as user-entered profile locations, gender or age differ from reality.  This in turn implies that user-generated profile content cannot be used as a useful proxy for the sociodemographic information~\cite{Graham2017}.

Many studies have overcome this limitation by taking advantage of the geolocation feature allowing Twitter users to include in their posts the location from which they were tweeted. Based on this metadata, studies have been able to assign home location to geolocated users with varying degrees of accuracy~\cite{Ajao2015}. Subsequent work has also been devoted to assigning to each user some indicator that might characterize their socioeconomic status based on their estimated home location. These indicators are generally extracted from other datasets used to complete the Twitter one, namely census data~\cite{Eagle, Eisenstein,EMoro} or real estate online services as Zillow.com~\cite{Zillow}. Other approaches have also relied on sources of socioeconomic information such as the UK Standard Occupation Classification (SOC) hierarchy, to assign socioeconomic status to users with occupation mentions ~\cite{Preot2015}. Despite the relative success of these methods, their common limitation is to provide observations and predictions based on a carefully hand-picked small set of users, letting alone the problem of socioeconomic status inference on larger and more heterogeneous populations. Our work stands out from this well-established line of research by expanding the definition of socioeconomic status to include several demographic features as well as by pinpointing potential home location to individual users with an unprecedented accuracy. Identifying socioeconomic status and the network effects of homophily\cite{Mishkovski2013} is an open question~\cite{Fixman2016}. However, recent results already showed that status homophily, i.e. the tendency of people of similar socioeconomic status are better connected among themselves, induce structural correlations which are pivotal to understand the stratified structure of society~\cite{Leo2016Socioeconomic}. While we verify the presence of status homophily in the Twitter social network, we detect further sociolinguistic correlations between language, location, socioeconomic status, and time, which may inform novel methods to infer socioeconomic status for a broader set of people using common information available on Twitter.

\section{Data Description}
\label{sec:data}

One of the main achievements of our study was the construction of a combined dataset for the analysis of sociolinguistic variables as a function of  socioeconomic status, geographic location, time, and the social network. As follows, we introduce the two aforementioned independent datasets and how they were combined. We also present a brief cross-correlation analysis to ground the validity of our combined dataset for the rest of the study. In what follows, it should also be noted that regression analysis was performed via linear regression as implemented in the Scikit Learn Toolkit  while data preprocessing and network study were performed using respectively pandas \cite{pandas} and NetworkX \cite{networkx} Python libraries.

\subsection{Twitter dataset: sociolinguistic features}

Our first dataset consists of a large data corpus collected from the online news and social networking service, Twitter. On it,  users can  post and interact with messages, "tweets", restricted to 140 characters. Tweets may come with several types of metadata including information about the author's profile, the detected language, where and when the tweet was posted, etc. Specifically, we recorded $170$ million tweets written in French, posted by $2.5$ million users in the timezones GMT and GMT+1 over three years (between July 2014 to May 2017). These tweets were obtained via the Twitter powertrack API feeds provided by Datasift and Gnip with an access rate varying between $15-25\%$\footnote{In order to uphold the strict privacy laws in France as well as the agreement signed with our data provider GNIP, full disclosure of the original dataset is not possible. Data collection and preprocessing pipelines could however be released upon request.}. 
\paragraph{Linguistic data:}
To obtain meaningful linguistic data we preprocessed the incoming tweet stream in several ways. As our central question here deals with the variability of the language, repeated tweets do not bring any additional information to our study. Therefore, as an initial filtering step, we decided to remove retweets. Next, in order to facilitate the detection of the selected  linguistic markers we removed any URLs, emoticons, mentions of other users (denoted by the \texttt{@} symbol) and hashtags (denoted by the \texttt{\#} symbol) from each tweet. These expressions were not considered to be semantically meaningful and their filtering allowed to further increase the speed and accuracy of our linguistic detection methods when run across the data. In addition we completed a last step of textual preprocessing by down-casing and stripping the punctuation out of the tweets body. POS-taggers such as MElt \cite{melt} were also tested but they provided no significant improvement in the detection of the linguistic markers.

\paragraph{Network data:}
We used the collected tweets in another way to infer social relationships between users. Tweet messages may be direct interactions between users, who mention each other in the text by using the \texttt{@} symbol (\texttt{@username}). When one user $u$, mentions another user $v$, user $v$ will see the tweet posted by user $u$ directly in his / her feed and may tweet back. In our work we took direct mentions as proxies of social interactions and used them to identify social ties between pairs of users. Opposite to the follower network, reflecting passive information exposure and less social involvement, the mutual mention network has been shown \cite{huberman} to capture better the underlying social structure between users. We thus use this network definition in our work as links are a greater proxy for social interactions.

In our definition we assumed a tie between users if they  mutually mentioned each other at least once during the observation period. People who reciprocally mentioned each other express some mutual interest, which may be a stronger reflection of real social relationships as compared to the non-mutual cases~\cite{Hadrien}. This constraint reduced the egocentric social network considerably leading to a directed structure of $508,975$ users and $4,029,862$ links that we considered being undirected in what follows.

\begin{figure*}[ht!]
\includegraphics[,width=.95\linewidth]{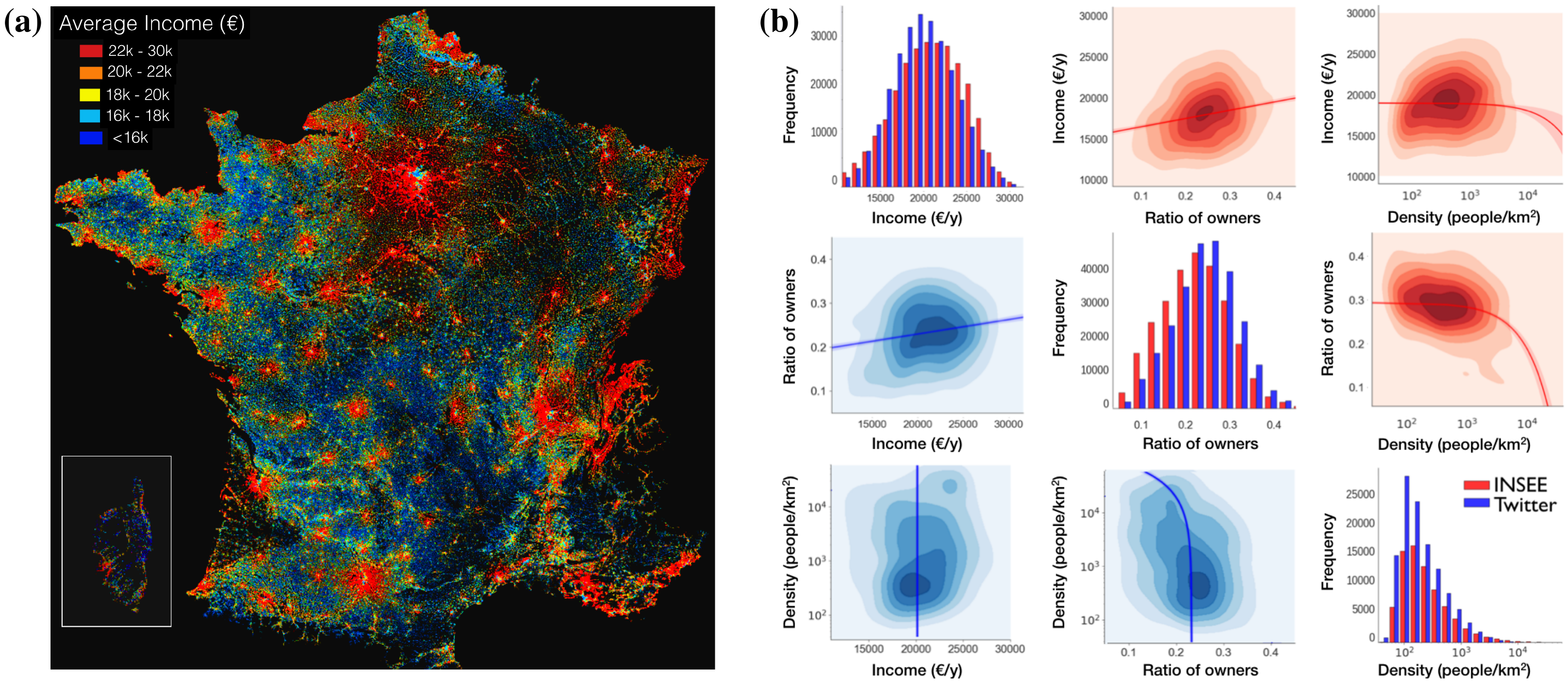}
\caption{Distributions and correlations of socioeconomic indicators. (a) Spatial distribution of average income in France with $200m \times 200m$ resolution. (b) Distribution of socioeconomic indicators (in the diag.) and their pairwise correlations measured in the INSEE (upper diag. panels) and Twitter geotagged (lower diag. panels) datasets. Contour plots assign the equidensity lines of the scatter plots, while solid lines are the corresponding linear regression values. Population density in $\log$.}
\label{fig:1}
\end{figure*}

\paragraph{Geolocated data:}
About $2\%$ of tweets included in our dataset contained some location information regarding either the tweet author's self-provided position or the place from which the tweet was posted. These pieces of information appeared as the combination of self reported locations or usual places tagged with GPS coordinates at different geographic resolution. We considered only tweets which contained the exact GPS coordinates with resolution of $\sim 3$ meters of the location where the actual tweet was posted. This actually means that we excluded tweets where the user assigned a place name such as "Paris" or "France" to the location field, which are by default associated to the geographical center of the tagged areas. Practically, we discarded coordinates that appeared more than $500$ times throughout the whole GPS-tagged data, assuming that there is no such $3\times 3$ meter rectangle in the country where $500$ users could appear and tweet by chance. After this selection procedure we rounded up each tweet location to a $100$ meter precision.

To obtain a unique representative location of each user, we extracted the sequence of all declared locations from their geolocated tweets. Using this set of locations we selected the most frequent to be the representative one, and we took it as a proxy for the user's home location. Further we limited our users to ones located throughout the French territory thus not considering others tweeting from places outside the country. This selection method provided us with $110,369$ geolocated users who are either detected as French speakers or assigned to be such by Twitter and all associated to specific 'home' GPS coordinates in France. To verify the spatial distribution of the selected population, we further assessed the correlations between the true population distributions (obtained from census data~\cite{INSEEpop}) at different administrative level and the geolocated user distribution aggregated correspondingly. More precisely, we computed the $R^2$ coefficient of variation between the inferred and official population distributions (a) at the level of $22$ regions\footnote{Note that since $2016$ France law determines $13$ metropolitan regions, however the available data shared by INSEE~\cite{INSEEpop} contained information about the earlier administrative structure containing $22$ regions.}. Correlations at this level induced a high coefficient of $R^2\simeq 0.89$ ($p<10^{-2}$); (b) At the arrondissement level with $322$ administrative units and coefficient $R^2\simeq 0.87$ ($p<10^{-2}$); and (c) at the canton level with $4055$ units with a coefficient $R\simeq 0.16$ ($p<10^{-2}$). Note that the relatively small coefficient at this level is due to the interplay of the sparsity of the inferred data and the fine grained spatial resolution of cantons. All in all, we can conclude that our sample is highly representative in terms of spatial population distribution, which at the same time validate our selection method despite the potential inherent biases induced by the method taking the most frequented GPS coordinates as the user's home location.

\subsection{INSEE dataset: socioeconomic features}
\label{sec:INSEE}

The second dataset we used was released in December 2016 by the National Institute of Statistics and Economic Studies (INSEE) of France. This data corpus~\cite{INSEEdata} contains a set of sociodemographic aggregated indicators, estimated from the 2010 tax return in France, for each 4 hectare ($200m \times 200m$) square patch across the whole French territory. Using these indicators, one can estimate the distribution of the average socioeconomic status (SES) of people with high spatial resolution. In this study, we concentrated on three indicators for each patch $i$ , which we took to be good proxies of the socioeconomic status of the people living within them. These were the $S^i_\mathrm{inc}$ average yearly income per capita (in euros), the $S^i_{\mathrm{own}}$ fraction of owners (not renters) of real estate, and the $S^i_\mathrm{den}$ density of population defined respectively as\begin{equation}:
S^i_\mathrm{inc}=\frac{{S}^i_{hh}}{{N}^i_{hh}}, \hspace{.15in} S^i_\mathrm{own}=\frac{N^i_\mathrm{own}}{N^i}, \hspace{.15in}\mbox{and}\hspace{.15in}  S^i_\mathrm{den}=\frac{N^i}{(200m)^2}.
\end{equation}
Here ${S}^i_{hh}$ and ${N}^i_{hh}$ assign respectively the cumulative income and total number of inhabitants of patch $i$, while $N^i_\mathrm{own}$ and $N^i$ are respectively the number of real estate owners  and the number of individuals living in patch $i$. As an illustration we show the spatial distribution of $S^i_\mathrm{inc}$ average income over the country in Fig.\ref{fig:1}a.

In order to uphold current privacy laws and due to the highly sensitive nature of the disclosed data, some statistical pretreatments were applied to the data by INSEE before its public release. More precisely, neighboring patches with less than $11$ households were merged together, while some of the sociodemographic indicators were winsorized. This set of treatments induced an inherent bias responsible for the deviation of the distribution of some of the socioeconomic indicators. These quantities were expected to be determined by the Pareto principle, thus reflecting the high level of socioeconomic imbalances present within the population. Instead, as shown in Fig.\ref{fig:1}b [diagonal panels], distributions of the derived socioeconomic indicators (in blue) appeared somewhat more symmetric than expected. This doesn't hold though for $P(S^i_\mathrm{den})$ (shown on a log-log scale in the lowest right panel of Fig.\ref{fig:1}b), which emerged with a broad tail similar to an expected power-law Pareto distribution. In addition, although the patches are relatively small ($200m \times 200m$), the socioeconomic status of people living may have some local variance, what we cannot consider here. Nevertheless, all things considered, this dataset and the derived socioeconomic indicators yield the most fine-grained description, allowed by national law, about the population of France over its whole territory. 

Despite the inherent biases of the selected socioeconomic indicators, in general we found weak but significant pairwise correlations between these three variables as shown in the upper diagonal panels in Fig.\ref{fig:1}b (in red), with values in Table \ref{tab:1}. We observed that while $S_\mathrm{inc}^{i}$ income and $S_\mathrm{own}^{i}$ owner ratio are positively correlated ($R=0.24$, $p<10^{-2}$), and the $S_\mathrm{own}^{i}$ and $S_\mathrm{den}^{i}$ population density are negatively correlated ($R=-0.23$, $p<10^{-2}$),  $S_\mathrm{inc}^{i}$ and $S_\mathrm{den}^{i}$ appeared to be very weakly correlated ($R=-0.07$, $p<10^{-2}$). This nevertheless suggested that high average income, high owner ratio, and low population density are consistently indicative of high socioeconomic status in the dataset.
\captionsetup[subfigure]{justification=justified,singlelinecheck=false}
\begin{table}[h!]
\centering
\caption{Pearson correlations and $p$-values measured between SES indicators in the INSEE and Twitter datasets.}
\label{tab:1}
\begin{tabular}{ cccc } 
 \hline
   & $S^i_\mathrm{inc} \sim S^i_\mathrm{own}$ & $S^i_\mathrm{inc} \sim S^i_\mathrm{den}$ & $S^i_\mathrm{own} \sim S^i_\mathrm{den}$ \\ \midrule
 INSEE & $0.24$ ($p<10^{-2}$) & $-0.07$ ($p<10^{-2}$)  & $-0.23$ ($p<10^{-2}$)  \\ 
 Twitter & $0.19$ ($p<10^{-2}$) & $0.00$ ($p>10^{-2}$) & $-0.22$ ($p<10^{-2}$) \\ 
 \bottomrule
\end{tabular}
\end{table}
\subsection{Combined dataset: individual socioeconomic features}

Data collected from Twitter provides a large variety of information about several users including their tweets, which disclose their interests, vocabulary, and linguistic patterns; their direct mentions from which their social interactions can be inferred; and the sequence of their locations, which can be used to infer their representative location. However, no information is directly available regarding their socioeconomic status, which can be pivotal to understand the dynamics and structure of their personal linguistic patterns. 

To overcome this limitation we combined our Twitter data with the socioeconomic maps of INSEE by assigning each geolocated Twitter user to a patch closest to their estimated home location (within 1 km).   This way we obtained for all $110,369$ geolocated users their dynamical linguistic data, their egocentric social network as well as a set of SES indicators. 

Such a dataset associating language with socioeconomic status and social network throughout the French metropolitan territory is unique to our knowledge and provides unrivaled opportunities to verify sociolinguistic patterns observed over a long period on a small-scale, but never established in such a large population.

To verify whether the geolocated Twitter users yet provide a representative sample of the whole population we compared the distribution and correlations of the their SES indicators to the population measures. Results are shown in Fig.\ref{fig:1}b diagonal (red distributions) and lower diagonal panels (in blue) with correlation coefficients and $p$-values summarized in Table.\ref{tab:1}. Even if we observed some discrepancy between the corresponding distributions and somewhat weaker correlations between the SES indicators, we found the same significant correlation trends (with the exception of the pair density / income) as the ones seen when studying the whole population, assuring us that each indicator correctly reflected the SES of individuals. 

\section{Linguistic variables}

We identified the following three linguistic markers  to study across users from different socioeconomic backgrounds: Correlation with SES has been evidenced for all of them. The optional deletion of negation is typical of spoken French, whereas the omission of the mute letters marking the plural in the nominal phrase is a variable cue of French writing. The third linguistic variable is a global measure of the lexical diversity of the Twitter users. We present them here in greater detail.

\subsection{Standard usage of negation}

The basic form of negation in French includes two negative particles: \textit{ne} (no) before the verb and another particle after the verb that conveys more accurate meaning: \textit{pas} (not), \textit{jamais} (never), \textit{personne} (no one), \textit{rien} (nothing), etc. Due to this double construction, the first part of the negation (\textit{ne}) is optional in spoken French, but it is obligatory in standard writing.
Sociolinguistic studies have previously observed the realization of  \textit{ne} in corpora of recorded everyday spoken interactions. Although all the studies do not converge, a general trend is that \textit{ne} realization is more frequent in speakers with higher socioeconomic status than in speakers with lower status~\cite{Ashby2001,Hansen2004}. We built upon this research to set out to detect both negation variants in the tweets using regular expressions.\footnote{Negation:\texttt{$\backslash \backslash \text{b}(\text{pas}\textbar \text{pa}\textbar \text{aps}\textbar \text{jamais}\textbar \text{ni}\textbar \text{personne}\textbar \text{rien}\textbar \text{ri1}\textbar \text{r1}\textbar \text{aucun}\textbar \text{aucune})\backslash \backslash \text{b}$}\\ \hspace{1cm}Standard Negation:\texttt{$\^{o}.*\backslash \backslash \text{b}(\text{ne}\textbar \text{n'})\backslash \backslash \text{b}.*\backslash \$$}} We are namely interested in the rate of usage of the standard negation (featuring both negative particles) across users:\\ 
\begin{equation}
L^u_{\mathrm{cn}}=\frac{n^u_{\mathrm{cn}}}{n^u_{\mathrm{cn}}+n^u_{\mathrm{incn}}} \hspace{.2in} \mbox{and} \hspace{.2in} \overline{L}^{i}_{\mathrm{cn}}=\frac{\sum_{u\in i}L^u_{\mathrm{cn}}}{N_i},
\end{equation}
where $n^{u}_{\mathrm{cn}}$ and $n^{u}_{\mathrm{incn}}$ assign the number of correct negation and incorrect number of negation of user $u$, thus $L_{\mathrm{cn}}^u$ defines the rate of correct negation of a users and $\overline{L}_{\mathrm{cn}}^i$ its average over a selected $i$ group (like people living in a given place) of $N_i$ users.

\subsection{Standard usage of plural ending of written words}

In written French, adjectives and nouns are marked as being plural by generally adding the letters \textit{s} or \textit{x} at the end of the word. Because these endings are mute (without counterpart in spoken French), their omission is the most frequent spelling error in adults~\cite{LucieMillet}. Moreover, studies showed correlations between standard spelling and social status of the writers, in preteens, teens and adults~\cite{brissaud,LucieMillet,Totereau2014}. We then set to estimate the use of standard plural across users: 
\begin{equation}
L^u_{\mathrm{cp}}=\frac{n^u_{\mathrm{cp}}}{n^u_{\mathrm{cp}}+n^u_{\mathrm{incp}}} \hspace{.2in} \mbox{and} \hspace{.2in} \overline{L}^{i}_{\mathrm{cp}}=\frac{\sum_{u\in i}L^u_{\mathrm{cp}}}{N_i}
\end{equation}
where the notation follows as before ($\mathrm{cp}$ stands for correct plural and $\mathrm{incp}$ stands for incorrect plural).

\subsection{Normalized vocabulary set size}

A positive relationship between an adult's lexical diversity level and his or her socioeconomic status has been evidenced in the field of language acquisition. Specifically, converging results showed that the growth of child lexicon depends on the lexical diversity in the speech of the caretakers, which in turn is related to their socioeconomic status and their educational level~\cite{Hoff2003,Hutten2007}. We thus proceeded to study the following metric: 
\begin{equation}
L^u_\mathrm{vs}=\frac{N^u_\mathrm{vs}}{N^u_{tw}} \hspace{.2in} \mbox{and} \hspace{.2in} \overline{L}^{i}_\mathrm{vs}=\frac{\sum_{u\in i}N^u_\mathrm{vs}}{N_i},
\end{equation}
where $N_vs^u$ assigns the total number of unique words used by user $u$ who tweeted $N_{tw}^u$ times during the observation period. As such $L_\mathrm{vs}^u$ gives the normalized vocabulary set size of a user $u$, while $\overline{L}_\mathrm{vs}^i$ defines its average for a population $i$.

\section{Results}

By measuring the defined linguistic variables in the Twitter timeline of users we were finally set to address the core questions of our study, which dealt with linguistic variation. More precisely, we asked whether the language variants used online depend on the socioeconomic status of the users, on the location or time of usage, and on ones social network. To answer these questions we present here a multidimensional correlation study on a large set of Twitter geolocated users, to which we assigned a representative location, three SES indicators, and a set of meaningful social ties based on the collection of their tweets. 

\subsection{Socioeconomic variation}

The socioeconomic status of a person is arguably correlated with education level, income, habitual location, or even with ethnicity and political orientation and may strongly determine to some extent patterns of individual language usage. Such dependencies have been theoretically proposed before~\cite{Labov1966Social}, but have rarely been inspected at this scale yet. The use of our previously described datasets enabled us to do so via the measuring of correlations between the inferred SES indicators of Twitter users and the use of the previously described linguistic markers.

\begin{figure}[htb]
\includegraphics[,width=.95\linewidth]{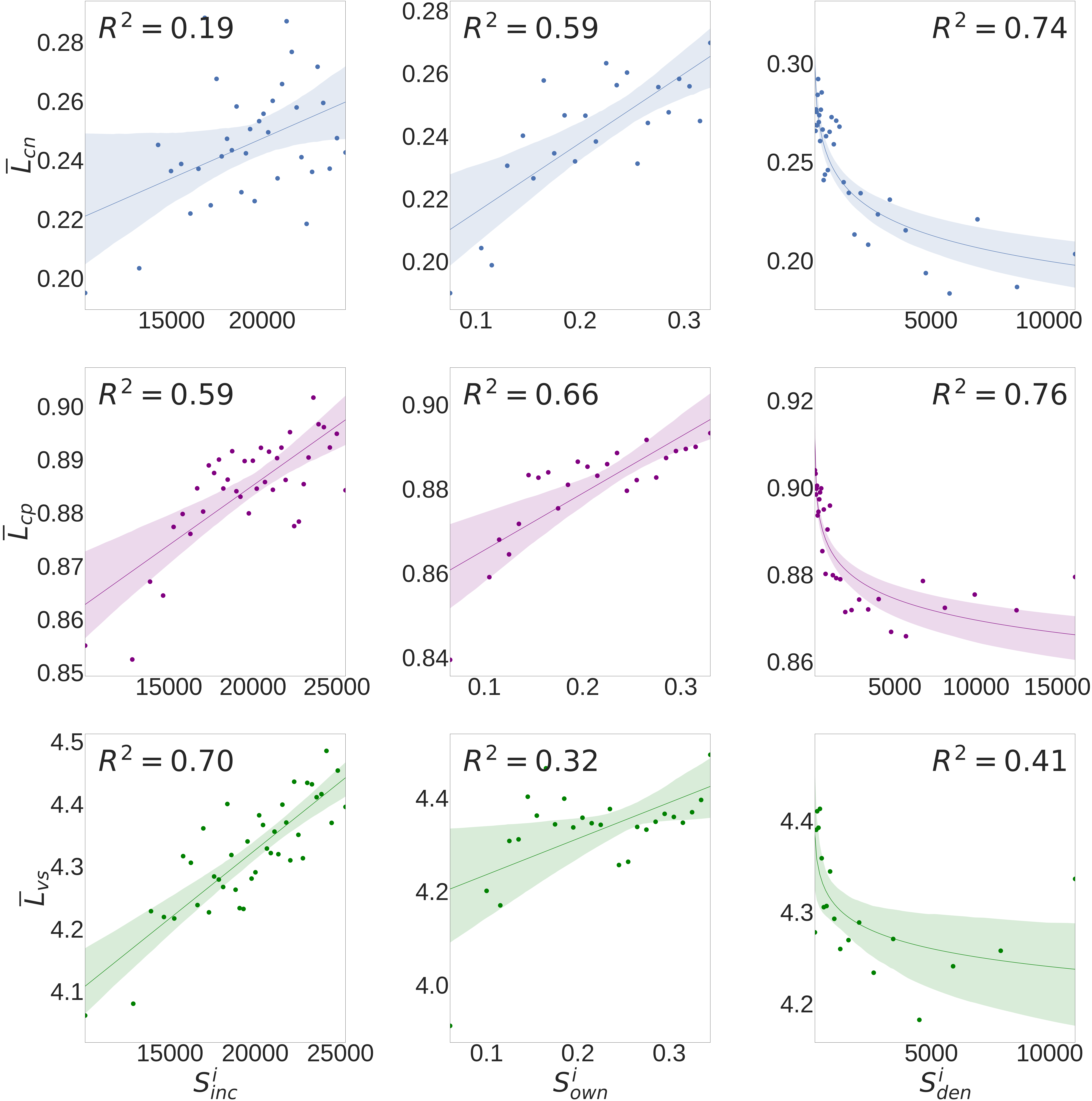}
\caption{Pairwise correlations between three SES indicators and three linguistic markers. Columns correspond to SES indicators (resp. $S^i_\mathrm{inc}$, $S^i_\mathrm{own}$, $S^i_\mathrm{den}$), while rows correspond to linguistic variables (resp. $\overline{L}_{\mathrm{cn}}$,  $\overline{L}_{\mathrm{cp}}$ and $\overline{L}_\mathrm{vs}$). On each plot colored symbols are binned data values and a linear regression curve are shown together with the $95$ percentile confidence interval and $R^2$ values.}
\label{fig:2}
\end{figure}

To compute and visualize these correlations we defined linear bins (in numbers varying from 20 to 50) for the socioeconomic indicators and computed the average of the given linguistic variables for people falling within the given bin. These binned values (shown as symbols in Fig.\ref{fig:2}) were used to compute linear regression curves and the corresponding confidence intervals (see Fig.\ref{fig:2}). An additional transformation was applied to the SES indicator describing population density, which was broadly distributed (as discussed in Section \ref{sec:INSEE} and Fig.\ref{fig:1}b), thus, for the regression process, the logarithm of its values were considered. To quantify pairwise correlations we computed the $R^2$ coefficient of determination values in each case.

\begin{table}[h!]
\centering
\caption{The $R^2$ coefficient of determination and the corresponding $p$-values computed for the pairwise correlations of SES indicators and linguistic variables.}
\label{tab:2}
\begin{tabular}{ c| ccc } 
 \hline
   & $S^i_\mathrm{inc}$ & $S^i_\mathrm{own}$ & $S^i_\mathrm{den}$ \\ \midrule
 $\overline{L}_{\mathrm{cn}}$  & $0.19$ ($p<10^{-2}$) & $0.59$ ($p<10^{-2}$)  & $0.74$ ($p<10^{-2}$)  \\ 
 $\overline{L}_{\mathrm{cp}}$ & $0.59$ ($p<10^{-2}$) & $0.66$ ($p<10^{-2}$) & $0.76$ ($p<10^{-2}$) \\ 
 $\overline{L}_\mathrm{vs}$ & $0.70$ ($p<10^{-2}$) & $0.32$ ($p<10^{-2}$) & $0.41$ ($p<10^{-2}$) \\ 
  \bottomrule
\end{tabular}
\end{table}

In Fig.\ref{fig:2} we show the correlation plots of all nine pairs of SES indicators and linguistic variables together with the linear regression curves, the corresponding $R^2$ values and the $95$ percentile confidence intervals (note that all values are also in Table \ref{tab:2}). These results show that correlations between socioeconomic indicators and linguistic variables actually exist. Furthermore, these correlation trends suggest that people with lower SES may use more non-standard expressions (higher rates of incorrect negation and plural forms) have a smaller vocabulary set size than people with higher SES. Note that, although the observed variation of linguistic variables were limited, all the correlations were statistically significant ($p<10^{-2}$) with considerably high $R^2$ values ranging from $0.19$ (between $\overline{L}_{\mathrm{cn}}\sim S_\mathrm{inc}$) to $0.76$ (between $\overline{L}_{\mathrm{cp}}\sim S_\mathrm{den}$). For the rates of standard negation and plural terms the population density appeared to be the most determinant indicator with $R^2=0.74$ (and $0.76$ respectively), while for the vocabulary set size the average income provided the highest correlation (with $R^2=0.7$).\\ One must also acknowledge that while these correlations exhibit high values consistently across linguistic and socioeconomic indicators, they only hold meaning at the population level at which the binning was performed. When the data is considered at the user level, the variability of individual language usage hinders the observation of the aforementioned correlation values (as demonstrated by the raw scatter plots (grey symbols) in Fig.~\ref{fig:2}).

\subsection{Spatial variation}

Next we chose to focus on the spatial variation of linguistic variables. Although officially a standard language is used over the whole country, geographic variations of the former may exist due to several reasons ~\cite{perozzi_freshman,wieling}. For instance, regional variability resulting from remnants of local languages that have disappeared, uneven spatial distribution of socioeconomic potentials, or influence spreading from neighboring countries might play a part in this process. For the observation of such variability, by using their representative locations, we assigned each user to a department of France. We then computed the $\overline{L}^{i}_{\mathrm{cn}}$ (resp. $\overline{L}^{i}_{\mathrm{cp}}$) average rates of standard negation (resp. plural agreement) and the $\overline{L}^{i}_\mathrm{vs}$ average vocabulary set size for each "d\'epartement" $i$  in the country (administrative division of France -- There are 97 d\'epartements).

\begin{figure}[htb]
\centering
\includegraphics[width=.95\linewidth]{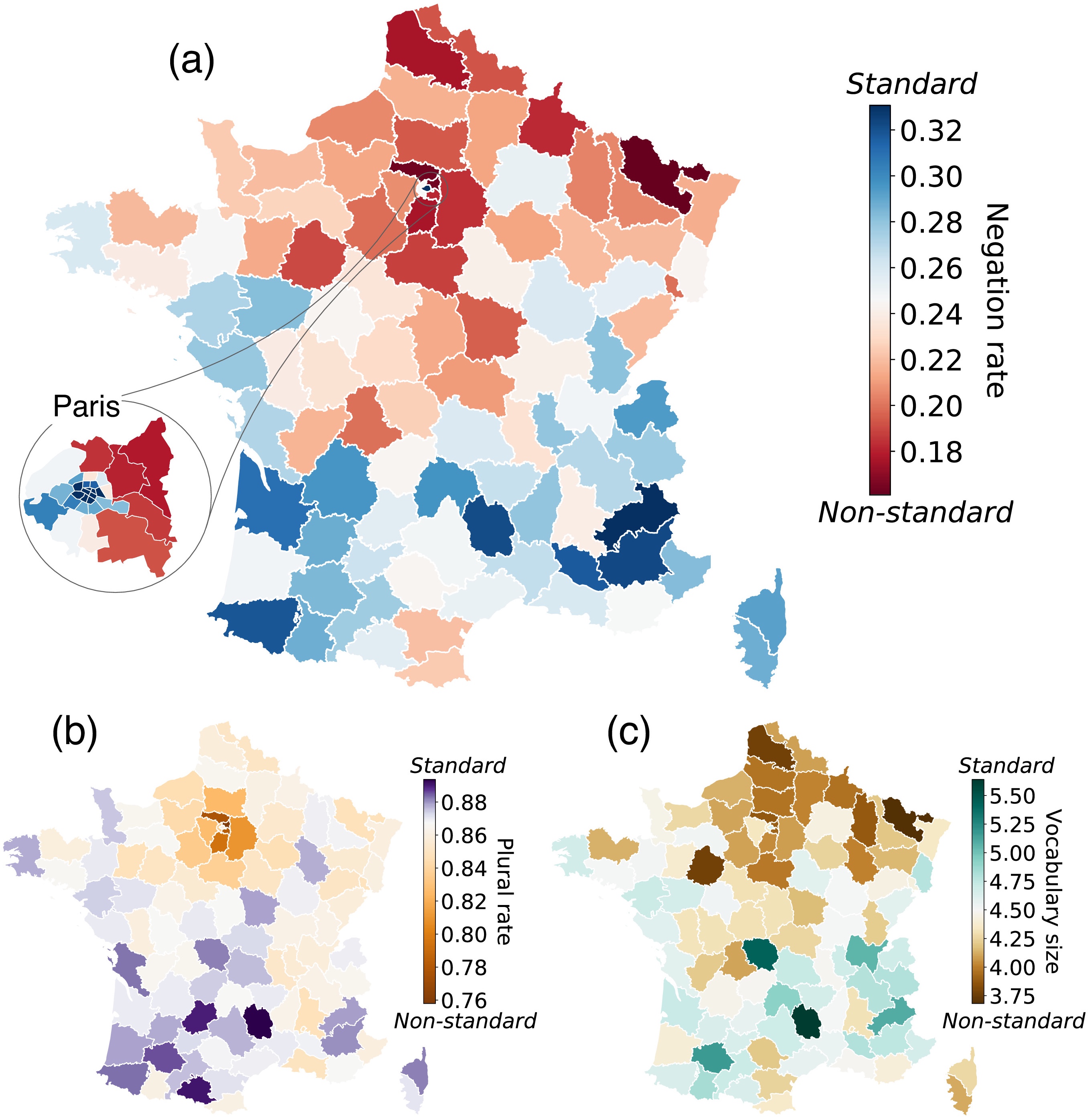}
\vspace{-0.1cm}
\caption{Geographical variability of linguistic markers in France. (a) Variability of the rate of correct negation. Inset focuses on larger Paris. (b) Variability of the rate of correct plural terms. (c) Variability of the average vocabulary size set. Each plot depicts variability on the department level except the inset of (a) which is on the "arrondissements" level.}
\label{fig:3}
\end{figure}

Results shown in Fig.\ref{fig:3}a-c revealed some surprising patterns, which appeared to be consistent for each linguistic variable. By considering latitudinal variability it appeared that, overall, people living in the northern part of the country used a less standard language, i.e., negated and pluralized less standardly, and used a smaller number of words. On the other hand, people from the South used a language which is somewhat closer to the standard (in terms of the aforementioned linguistic markers) and a more diverse vocabulary. The most notable exception is Paris, where in the city center people used more standard language, while the contrary is true for the suburbs. This observation, better shown in Fig.\ref{fig:3}a inset, can be explained by the large differences in average socioeconomic status between districts. Such segregation is known to divide the Eastern and Western sides of suburban Paris, and in turn to induce apparent geographic patterns of standard language usage. We found less evident longitudinal dependencies of the observed variables. Although each variable shows a somewhat diagonal trend, the most evident longitudinal dependency appeared for the average rate of standard pluralization (see Fig.\ref{fig:3}b), where users from the Eastern side of the country used the language in less standard ways. Note that we also performed a multivariate regression analysis (not shown here), using the linguistic markers as target and considering as factors both location (in terms of latitude and longitude) as and income as proxy of socioeconomic status. It showed that while location is a strong global determinant of language variability, socioeconomic variability may still be significant locally to determine standard language usage (just as we demonstrated in the case of Paris).

\subsection{Temporal variation}

Another potentially important factor determining language variability is the time of day when users are active in Twitter \cite{hamilton_diachronic,stat_perozzi}. The temporal variability of standard language usage can be measured for a dynamical quantity like the $L_{\mathrm{cn}}(t)$ rate of correct negation. To observe its periodic variability (with a $\Delta T$ period of one week) over an observation period of $T$ (in our case $734$ days), we computed 
\begin{equation}
\overline{L}^{\Lambda}_{\mathrm{cn}}(t)=\frac{\Delta T}{|\Lambda|T}\sum_{u\in \Lambda}\sum_{k=0}^{\left \lfloor{T/\Delta T}\right \rfloor }L_{\mathrm{cn}}^{u}(t+k\Delta T),
\end{equation}
in a population $\Lambda$ of size $|\Lambda|$ with a time resolution of one hour. This quantity reflects the average standard negation rate in an hour over the week in the population $\Lambda$. Note that an equivalent $\overline{L}^{\Lambda}_{\mathrm{cp}}(t)$ measure can be defined for the rate of standard plural terms, but not for the vocabulary set size as it is a static variable.

\begin{figure}[t]
\centering
 \includegraphics[width=.95\linewidth]{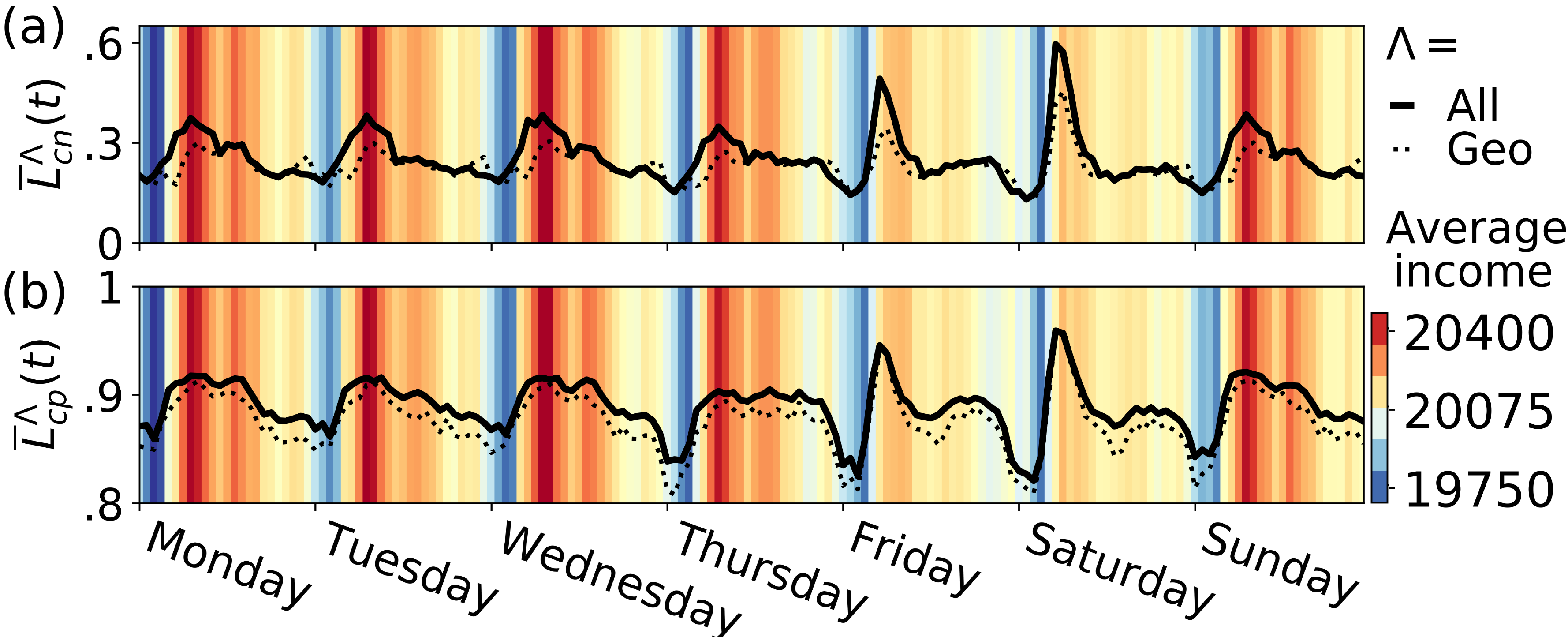}
 \vspace{-0.2cm}
\caption{Temporal variability of (a) $\overline{L}^{\Lambda}_{\mathrm{cn}}(t)$ (resp. (b) $\overline{L}^{\Lambda}_{\mathrm{cp}}(t)$) average rate of correct negation (resp. plural terms) over a week with one hour resolution. Rates were computed for $\Lambda=all$ (solid line) and $\Lambda=geo$located Twitter users. Colors indicates the temporal variability of the average income of geolocated population active in a given hour.}
\label{fig:4}
\end{figure}

In Fig.~\ref{fig:4}a and b we show the temporal variability of $\overline{L}^{\Lambda}_{\mathrm{cn}}(t)$ and $\overline{L}^{\Lambda}_{\mathrm{cp}}(t)$ (respectively) computed for the whole Twitter user set ($\Gamma=all$, solid line) and for geolocated users ($\Gamma=geo$, dashed lines). Not surprisingly, these two curves were strongly correlated as indicated by the high Pearson correlation coefficients summarized in the last column of Table \ref{tab:3} which, again, assured us that our geolocated sample of Twitter users was representative of the whole set of users. At the same time, the temporal variability of these curves suggested that people tweeting during the day used a more standard language than those users who are more active during the night. However, after measuring the average income of active users in a given hour over a week, we obtained an even more sophisticated picture. It turned out that people active during the day have higher average income (warmer colors in Fig.~\ref{fig:4}) than people active during the night (colder colors in Fig.~\ref{fig:4}). Thus the variability of standard language patterns was largely explained by the changing overall composition of active Twitter users during different times of day and the positive correlation between socioeconomic status and the usage of higher linguistic standards (that we have seen earlier). This explanation was supported by the high coefficients (summarized in Table \ref{tab:3}), which were indicative of strong and significant correlations between the temporal variability of average linguistic variables and average income of the active population on Twitter.

\begin{table}[h!]
\centering
\caption{Pearson correlations and $p$-values of pairwise correlations of time varying $S_\mathrm{inc}(t)$ average income with $\overline{L}^{\Lambda}_{\mathrm{cn}}(t)$ and $\overline{L}^{\Lambda}_{\mathrm{cp}}(t)$ average linguistic variables; and between average linguistic variables of $\Lambda=all$ and $\Lambda=\text{geo}$-localized users.}
\footnotesize
\label{tab:3}
\hspace{-0.2cm}
\begin{tabular}{ c| ccc } 
 \hline
   & $\overline{L}^{all}_{*}(t)\sim S_\mathrm{inc}(t)$ & $\overline{L}^{geo}_{*}(t)\sim S_\mathrm{inc}(t)$ & $\overline{L}^{geo}_{*}(t)\sim \overline{L}^{all}_{*}(t)$ \\ \midrule
 $*={\mathrm{\tiny{cn}}}$ & $0.5915$ ($p<10^{-2}$) & $0.622$ ($p<10^{-2}$) & $0.805$ ($p<10^{-2}$) \\ 
 $*={\mathrm{\tiny{cp}}}$  & $0.7027$ ($p<10^{-2}$) & $0.665$ ($p<10^{-2}$) & $0.98021$ ($p<10^{-2}$) \\ 
 \bottomrule
\end{tabular}\vspace{-0.3cm}

\end{table}
\subsection{Network variation}

Finally we sought to understand the effect of the social network on the variability of linguistic patterns. People in a social structure can be connected due to several reasons. Link creation mechanisms like focal or cyclic closure~\cite{Kumpula2007Emergence, Laurent2015From}, or preferential attachment~\cite{Kunegis2013Preferential} together with the effects of homophily~\cite{Miller2001Birds} are all potentially driving the creation of social ties and communities, and the emergence of community rich complex structure within social networks. In terms of homophily, one can identify several individual characteristics like age, gender, common interest or political opinion, etc., that might increase the likelihood of creating relationships between disconnected but similar people, who in turn influence each other and become even more similar. Status homophily between people of similar socioeconomic status has been shown to be important~\cite{Leo2016Socioeconomic} in determining the creation of social ties and to explain the stratified structure of society. By using our combined datasets, we aim here to identify the effects of status homophily and to distinguish them from other homophilic correlations and the effects of social influence inducing similarities among already connected people.

To do so, first we took the geolocated Twitter users in France and partitioned them into nine socioeconomic classes using their inferred income $S_\mathrm{inc}^u$. Partitioning was done first by sorting users by their $S^u_\mathrm{inc}$ income to calculate their $C(S^u_\mathrm{inc})$ cumulative income distribution function. We defined socioeconomic classes by segmenting $C(S^u_\mathrm{inc})$ such that the sum of income is the same for each classes (for an illustration of our method see Fig.\ref{fig:A1}a in the Appendix). We constructed a social network by considering mutual mention links between these users (as introduced in Section \ref{sec:data}). Taking the assigned socioeconomic classes of connected individuals, we confirmed the effects of status homophily in the Twitter mention network by computing the connection matrix of socioeconomic groups normalized by the equivalent matrix of corresponding configuration model networks, which conserved all network properties except structural correlations (as explained in the Appendix). The diagonal component in Fig.\ref{fig:A1} matrix indicated that users of similar socioeconomic classes were better connected, while people from classes far apart were less connected than one would expect by chance from the reference model with users connected randomly.

\begin{figure}[ht!]
\centering
\includegraphics[width=\linewidth]{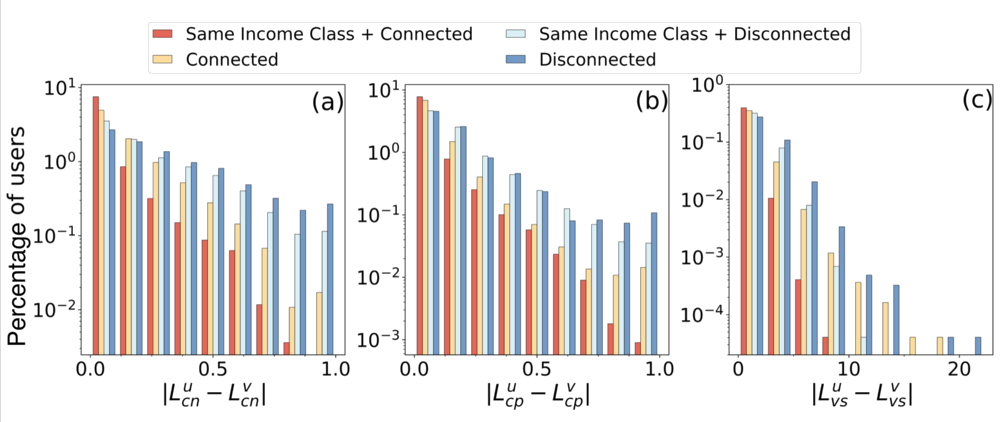}
\setlength{\belowcaptionskip}{-12pt}
\vspace{-0.25in}
\caption{Distribution of the $|L^{u}_{*}-L^{v}_{*}|$ absolute difference of linguistic variables $*\in \{\mathrm{cn},\mathrm{cp},vs\}$ (resp. panels (a), (b), and (c)) of user pairs who were connected and from the same socioeconomic group (red), connected (yellow), disconnected and from the same socioeconomic group (light blue), disconnected pairs of randomly selected users (blue).}
\label{fig:5}
\end{figure}

In order to measure linguistic similarities between a pair of users $u$ and $v$, we simply computed the $|L^{u}_{*}-L^{v}_{*}|$ absolute difference of their corresponding individual linguistic variable $*\in \{\mathrm{cn},\mathrm{cp},vs\}$. This measure appeared with a minimum of $0$ and associated smaller values to more similar pairs of users. To identify the effects of status homophily and the social network, we proceeded by computing the similarity distribution in four cases: for connected users from the same socioeconomic class; for disconnected randomly selected  pairs of users from the same socioeconomic class; for connected users in the network; and randomly selected pairs of disconnected users in the network. Note that in each case the same number of user pairs were sampled from the network to obtain comparable averages. This number was naturally limited by the number of connected users in the smallest socioeconomic class, and were chosen to be $10,000$ in each cases. By comparing the distributions shown in Fig.\ref{fig:5} we concluded that (a) connected users (red and yellow bars) were the most similar in terms of any linguistic marker. This similarity was even greater when the considered tie was connecting people from the same socioeconomic group; (b) network effects can be quantified by comparing the most similar connected (red bar) and disconnected (light blue bar) users from the same socioeconomic group. Since the similarity between disconnected users here is purely induced by status homophily, the difference of these two bars indicates additional effects  that cannot be explained solely by status homophily. These additional similarities may rather be induced by other factors such as social influence, the physical proximity of users within a geographical area or other homophilic effects that were not accounted for. (c) Randomly selected pairs of users were more dissimilar than connected ones as they dominated the distributions for larger absolute difference values. We therefore concluded that both the effects of network and status homophily mattered in terms of linguistic similarity between users of this social media platform.

\section{Conclusions}

The overall goal of our study was to explore the dependencies of linguistic variables on the socioeconomic status, location, time varying activity, and social network of users. To do so we constructed a combined dataset from a large Twitter data corpus, including geotagged posts and proxy social interactions of millions of users, as well as a detailed socioeconomic map describing average socioeconomic indicators with a high spatial resolution in France. The combination of these datasets provided us with a large set of Twitter users all assigned to their Twitter timeline over three years, their location, three individual socioeconomic indicators, and a set of meaningful social ties. Three linguistic variables extracted from individual Twitter timelines were then studied as a function of the former, namely, the rate of standard negation, the rate of plural agreement and the size of vocabulary set.

Via a detailed multidimensional correlation study we concluded that (a) socioeconomic indicators and linguistic variables are significantly correlated. i.e. people with higher socioeconomic status are more prone to use more standard variants of language and a larger vocabulary set, while people on the other end of the socioeconomic spectrum tend to use more non-standard terms and, on average, a smaller vocabulary set; (b) Spatial position was also found to be a key feature of standard language use as, overall, people from the North tended to use more non-standard terms and a smaller vocabulary set compared to people from the South; a more fine-grained analysis reveals that the spatial variability of language is determined to a greater extent locally by the socioeconomic status; (c) In terms of temporal activity, standard language was more likely to be used during the daytime while non-standard variants were predominant during the night. We explained this temporal variability by the turnover of population with different socioeconomic status active during night and day; Finally (d) we showed that the social network and status homophily mattered in terms of linguistic similarity between peers, as connected users with the same socioeconomic status appeared to be the most similar, while disconnected people were found to be the most dissimilar in terms of their individual use of the aforementioned linguistic markers.

Despite these findings, one has to acknowledge the multiple limitations affecting this work: First of all, although Twitter is a broadly adopted service in most technologically enabled societies, it commonly provides a biased sample in terms of age and socioeconomic status as older or poorer people may not have access to this technology. In addition, home locations inferred for lower activity users may induced some noise in our inference method. Nevertheless, we demonstrated that our selected Twitter users are quite representative in terms of spatial, temporal, and socioeconomic distributions once compared to census data. Other sources of bias include the "homogenization" performed by INSEE to ensure privacy rights are upheld as well as the proxies we devised to approximate users' home location and social network. Currently, a sample survey of our set of geolocated users is being conducted so as to bootstrap socioeconomic data to users and definitely validate our inference results.
Nonetheless, this INSEE dataset provides still the most comprehensive available information on socioeconomic status over the whole country. For limiting such risk of bias, we analyzed the potential effect of the confounding variables on distribution and cross-correlations of SES indicators. Acknowledging possible limitations of this study, we consider it as a necessary first step in analyzing income through social media using datasets orders of magnitude larger than in previous research efforts.

Finally we would like to emphasize two scientific merits of the paper. On one side, based on a very large sample, we confirm and clarify results from the field of sociolinguistics and we highlight new findings. We  thus confirm clear correlations between the variable realization of the negative particle in French and three indices of socioeconomic status. This result challenges those among the sociolinguistic studies that do not find such correlation. Our data also suggested that the language used in the southern part of France is more standard. Understanding this pattern fosters further investigations within sociolinguistics. We finally established that the linguistic similarity of socially connected people is partially explained by status homophily but could be potentially induced by social influences passing through the network of links or other terms of homophilic correlations. Beyond scientific merit, we can identify various straightforward applications of our results. The precise inference of socioeconomic status of individuals from online activities is for instance still an open question, which carries a huge potential in marketing design and other areas. Our results may be useful moving forward in this direction by using linguistic information, available on Twitter and other online platforms, to infer socioeconomic status of individuals from their position in the network as well as the way they use their language. 

\appendix
\section{Appendix: S\lowercase{tatus homophily}}
\vspace{-0.1cm}
\begin{figure}[ht!]
\centering
\includegraphics[width=.95\linewidth]{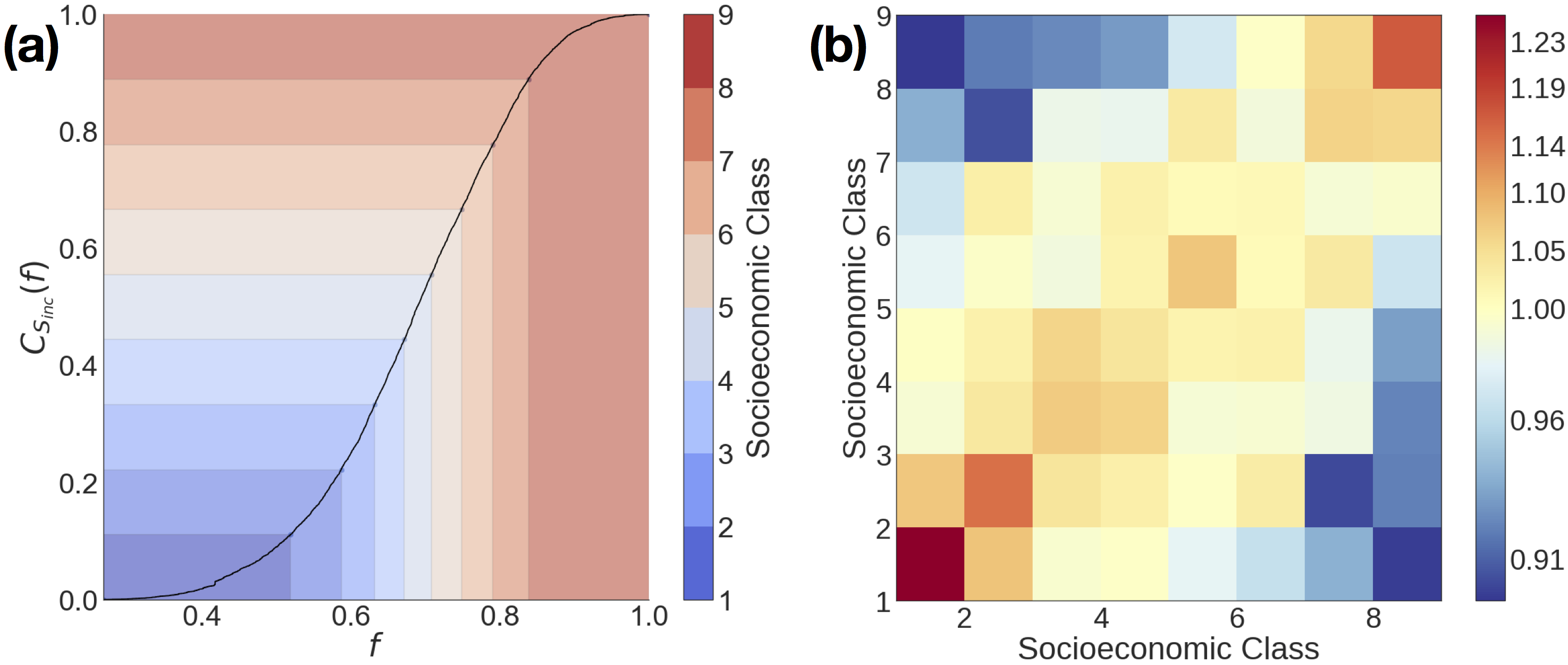}
\vspace{-0.2cm}
\setlength{\belowcaptionskip}{-6pt}
\caption{(a) Definition of socioeconomic classes by partitioning users into nine groups with the same cumulative annual income. (b) Structural correlations between SES groups depicted as matrix of the ratio  $|E(s_i, s_j)|/|E_{rand}(s_i, s_j)|$ between the original and the average randomized mention network}
\label{fig:A1}
\end{figure}

Status homophily in social networks appears as an increased tendency for people from similar socioeconomic classes to be connected. This correlation can be identified by comparing likelihood of connectedness in the empirical network to a random network, which conserves all network properties except structural correlations. To do so, we took each $(s_i,s_j)$ pair of the nine SES class in the Twitter network and counted the number of links $|E(s_i, s_j)|$ connecting people in classes $s_i$ and $s_j$. As a reference system, we computed averages over $100$ corresponding configuration model network structures~\cite{newman2010networks}.
To signalize the effects of status homophily, we took the ratio $|E(s_i, s_j)|/|E_{rand}(s_i, s_j)|$ of the two matrices (shown in Fig.\ref{fig:A1}b). The diagonal component in Fig.\ref{fig:A1}b with values larger than $1$ showed that users of the same or similar socioeconomic class were better connected in the original structure than by chance, while the contrary was true for users from classes far apart (see blue off-diagonal components). To verify the statistical significance of this finding, we performed a $\chi^2$-test, which showed that the distribution of links in the original matrix was significantly different from the one of the average randomized matrix  ($p<10^{-5}$). This observation verified status homophily present in the Twitter mention network.



\end{document}